\def\year{2022}\relax
\documentclass[letterpaper]{article} % DO NOT CHANGE THIS
\usepackage{aaai22}  % DO NOT CHANGE THIS
\usepackage{times}  % DO NOT CHANGE THIS
\usepackage{helvet}  % DO NOT CHANGE THIS
\usepackage{courier}  % DO NOT CHANGE THIS
\usepackage[hyphens]{url}  % DO NOT CHANGE THIS
\usepackage{graphicx} % DO NOT CHANGE THIS
\usepackage{natbib}  % DO NOT CHANGE THIS AND DO NOT ADD ANY OPTIONS TO IT
\usepackage{caption} % DO NOT CHANGE THIS AND DO NOT ADD ANY OPTIONS TO IT
\DeclareCaptionStyle{ruled}{labelfont=normalfont,labelsep=colon,strut=off} % DO NOT CHANGE THIS
\usepackage{algorithm}
\usepackage{algorithmic}

\usepackage{booktabs}
\usepackage{multirow}
\newtheorem{definition}{Definition}
\usepackage{amsmath}
\usepackage{amssymb}
\usepackage{bm}

\def\bx{\bm{x}}

\usepackage{xcolor}

\usepackage{newfloat}
\usepackage{listings}
\floatstyle{ruled}
\newfloat{listing}{tb}{lst}{}
\floatname{listing}{Listing}
\title{Regional Adversarial Training for Better Robust Generalization}
\author {
    Chuanbiao Song\textsuperscript{\rm 1}\equalcontrib,
    Yanbo Fan\textsuperscript{\rm 2}\equalcontrib, 
    Yichen Yang\textsuperscript{\rm 1}\equalcontrib,\\
    Baoyuan Wu\textsuperscript{\rm 3,4}\thanks{Corresponding authors.},
    Yiming Li\textsuperscript{\rm 5}, %
    Zhifeng Li\textsuperscript{\rm 2},
    Kun He\textsuperscript{\rm 1}\footnotemark[2]
}
\affiliations{
    \textsuperscript{\rm 1} Huazhong University of Science and Technology\\
    \textsuperscript{\rm 2} Tencent\\% AI Lab\\
    \textsuperscript{\rm 3} The Chinese University of Hong Kong, Shenzhen\\
    \textsuperscript{\rm 4} Shenzhen Research Institute of Big Data\\    
    \textsuperscript{\rm 5} Tsinghua University \\
    cbsong@hust.edu.cn, fanyanbo0124@gmail.com, yangyc@hust.edu.cn, 
    wubaoyuan1987@gmail.com, li-ym18@mails.tsinghua.edu.cn, michaelzfli@tencent.com, brooklet60@hust.edu.cn
}

\usepackage{bibentry}
\begin{document}

\maketitle

\begin{abstract}
	Adversarial training (AT) has been demonstrated as one of the most promising defense methods against various adversarial attacks. 
	%However,
	To our knowledge, 
	existing AT-based methods usually train with the locally most adversarial perturbed points and treat all the perturbed points equally, %leading to weak
	which may lead to considerably weaker 
	adversarial robust generalization on test data. In this work, we introduce a new adversarial training framework that considers the diversity as well as characteristics of the perturbed points in the vicinity of benign samples. To realize the framework, we propose %a novel method termed Regional Adversarial Training (RAT)
	a Regional Adversarial Training (RAT) defense method 
	that first utilizes the attack path generated by the typical iterative attack method of projected gradient descent (PGD), and constructs an adversarial region based on the attack path. Then, RAT samples diverse perturbed training points efficiently inside this region, and utilizes a distance-aware label smoothing mechanism to capture our intuition that perturbed points at different locations should have different impact on the model performance. Extensive experiments on several benchmark datasets show that RAT consistently makes significant improvement on standard adversarial training (SAT), and exhibits better robust generalization. 
	
	%In this work, we introduce a novel defense method called Regional Adversarial Training (RAT) that considers the diversity as well as characteristics of the perturbed points in the vicinity of benign samples. RAT first utilizes the attack path generated by a typical iterative attack method of projected gradient descent (PGD), and constructs an adversarial region based on the attack path. 
\end{abstract}

\section{Introduction}
\label{sec:introduction}
Deep learning models have achieved substantial success %across 
on a wide variety
of computer vision tasks
%in the area ofcomputer vision
~\cite{he2016deep,lin2014microsoft,girshick2015fast}. %, including image classification~\cite{he2016deep}, segmentation~\cite{lin2014microsoft} and object detection~\cite{girshick2015fast}, etc. 
%Despite the great progress
However, recent studies have shown that deep learning models are inherently vulnerable to \textit{adversarial examples}~\cite{SzegedyZSBEGF13,GoodfellowSS14,PapernotMG16,PapernotMGJCS17}, i.e., injecting imperceptible but malicious perturbations into the clean input could drastically degrade the model performance.

% Due to the security threat, a wide range of works~\cite{MadryMSTV18, DhillonALBKKA18, ZhangYJXGJ19,liao2018defense, CohenRK19, lee2020adversarial} have been proposed to improve the adversarial robustness. 

%Such vulnerability has imposed serious threats to the safety and security critical applications, such as autonomous driving~\cite{AmodeiOSCSM16} and malware detection~\cite{CuiXCCWC18}. %and it is therefore necessary to develop robust deep learning models.

% Among existing methods, 
Adversarial training (AT), which incorporates adversarial examples into the training set, is one of the most effective defense methods to improve the adversarial robustness of deep learning models~\cite{MadryMSTV18,ZhangYJXGJ19,lee2020adversarial,AthalyeC018,LiLWZG19}. 
% Its key idea is to train with a local adversarial perturbed point around each benign data point, and such perturbed point is assigned the same one-hot class label as the corresponding benign data point for adversarial training.
%To generate the locally most adversarial perturbed points used for training, a typical adopted and commonly used method is the projected gradient descent (PGD) method~\cite{MadryMSTV18}.
However, AT-based defense methods usually suffer from the weak robust generalization due to the following two reasons. 
% We investigate in this direction and observe that most existing AT-based methods neither explore the diversity of the perturbed training points nor utilize their characteristics during the training. 
% On one hand, %for the diversity of the perturbed training points, 
Firstly, AT-based methods usually adopt the locally most adversarial perturbed point around each benign data point as the perturbed training point, whereas ignoring the diversity of the perturbed training points and increasing the risk of robust overfitting~\cite{rice2020overfitting}. 
% On the other hand, %for the difference of the perturbed training points,
Secondly, they treat all the perturbed points in the $\epsilon$-ball of the benign points equally, i.e., assigning such points the same one-hot class label of their benign counterparts, but consider neither their characteristics nor differences. 
Thereby, most existing AT-based methods tend to overfit on the specific attack used in the training and 
exhibit weak robust generalization for the same threat model with different settings~\cite{TramerKPGBM18,SongHWH19}.

%For instance, as a typical AT-based method, the standard adversarial training (SAT)~\cite{MadryMSTV18} utilizes %the PGD attack
%the projected gradient descent (PGD) attack~\cite{MadryMSTV18}
%to find the locally most adversarial perturbed points as the perturbed training points, and labels such points with the same one-hot class label of their benign counterparts. These conservative constraints of adversarial training lead to severe degradation on the standard accuracy and robust overfitting under the insufficient model capacity~\cite{MadryMSTV18, TsiprasSETM19,abs-2010-01736}.

\begin{figure}[t]
\begin{center}    			
\includegraphics[width=0.47\textwidth]{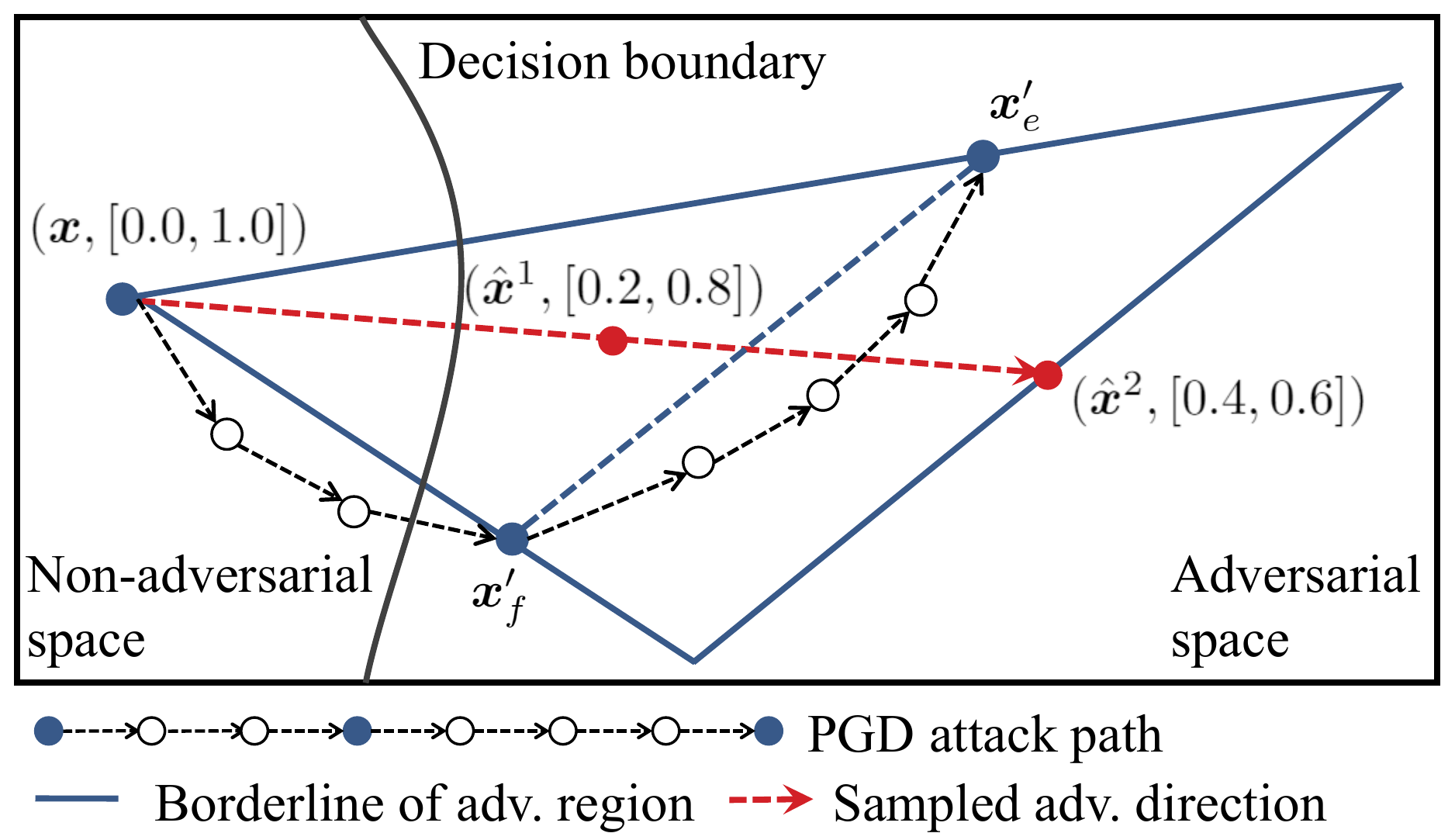}
\end{center}
\caption{Illustration of the adversarial region and the distance-aware labels for sampled training points. $\bx$ denotes the benign data point. $\bx'_f$ and $\bx'_e$ denote the first adversarial point and the end adversarial point along the attack path generated by iterative PGD method, respectively. $\hat{\bx}^1$ and $\hat{\bx}^2$ are sampled perturbed training points with different soft labels.
% \HK{$\hat{\bx}^1$ and $\hat{\bx}^2$ are sampled perturbed points with different soft labels.}
}
\label{fig: ball}
\vspace{-0.5em}	
\end{figure}

%In this paper, to alleviate the aforementioned issues, we propose a novel adversarial training method called the Regional Adversarial Training (RAT) that aims to enrich the perturbed training points with diverse magnitudes and diverse directions, and further treat these perturbed points differently based on their characteristics so as to improve the robust generalization.
To address the aforementioned issues, we propose a new adversarial training framework that advocates exploring the diversity of the perturbed training points and considering their characteristics and differences.
We realize this framework by designing a %novel defense method called the \textit{Regional Adversarial Training} (RAT).
 \textit{Regional Adversarial Training} (RAT) defense method.  
Firstly, RAT utilizes an \textit{Adversarial Region-based Sampler} (ARS) for sampling diverse perturbed training points. As illustrated in Figure~\ref{fig: ball}, the adversarial regional sampler first constructs an informative area with three points (the benign data point, the first adversarial point and the end adversarial point along the attack path generated by the typical iterative PGD attack~\cite{MadryMSTV18}), and enlarges the area in an appropriate proportion to obtain the adversarial region.
Then, sampling from the adversarial region can boost the diversity of the perturbed training points in terms of adversarial directions and magnitudes.
Secondly, RAT proposes a \textit{Distance-aware Label Smoothing} (DLS) mechanism to capture the intuition that perturbed training points at different locations should have different contributions to the model performance. 
Specifically, the distance-aware label smoothing mechanism aims to assign different soft labels for perturbed training points based on their characteristics, with the principle that a perturbed point closer to (farther from) the benign point should have higher (lower) confidence w.r.t. the ground-truth class label. 
%As shown in Figure~\ref{fig: ball}, with the distance between sampled perturbed point $\hat{\bx}^1$ and benign point $\bx$ shorter than that between sampled perturbed point $\hat{\bx}^2$ and $\bx$, the soft label we assign to $\hat{\bx}^1$ is closer to the true label of $\bx$.

%Treating the perturbed training points differently based on their characteristics could help the robust generalization under the insufficient model  capacity in adversarial training~\cite{abs-2010-01736}.

The main contributions of our work are as follows. %are threefold.
\begin{itemize}
\item We propose a new adversarial training framework that considers the diversity as well
as different characteristics of the perturbed training points in the vicinity of benign samples.
\item To realize the framework, we design %a novel defense method called the Regional Adversarial Training (RAT)
a Regional Adversarial Training (RAT) defense method 
that samples diverse perturbed training points in different directions and magnitudes, and assign different soft labels to these perturbed points based on their locations.
%utilizes the Adversarial Regional Sampler (ARS) for sampling diverse perturbed training points, and the Distance-aware Label Smoothing (DLS) for treating these perturbed points differently based on their locations. 
\item Extensive experiments on three standard datasets validate the superiority of RAT, which consistently makes significant improvements on the standard adversarial training without extra training cost and exhibits better robustness against various adversarial attacks. %standard and robust generalization compared to several state-of-the-art defenses.
\item Further analysis show that RAT brings smaller gaps on the standard and adversarial robust generalization, as compared to other AT-based methods, and RAT performs more stably with better robust generalization under the PGD attack with various settings. Moreover, RAT exhibits better robustness on natural image corruptions.

\end{itemize}

\section{Related Work}
\label{sec:relatedwork}

%We first provide a brief overview on adversarial attacks and defenses. 

Since the discovery of adversarial examples~\cite{GoodfellowSS14}, various adversarial attack methods have been proposed to deceive deep learning models with small but malicious perturbations.
In general, these attack methods can be categorized into two types, i.e., the {\it white-box attacks} and the {\it black-box attacks}, based on whether the attacker can access the inner information of the target model. Among which, the projected gradient descent method (PGD)~\cite{MadryMSTV18} is a typical white-box iterative attack method.
%White-box attacks include single-step attacks, e.g. fast gradient sign method (FGSM)~\cite{GoodfellowSS14} and iterative attacks, e.g. projected gradient descent method (PGD)~\cite{MadryMSTV18}. 
%Black-box attacks include transfer-based attacks~\cite{PapernotMGJCS17,LiuCLS17} and query-based attacks~\cite{LiLWZG19}.

Correspondingly, numerous methods have been proposed to defense against the adversarial attacks, %including 
which can be categorized into input transformation~\cite{buckman2018thermometer,YangZXK19}, model ensemble~\cite{PangXDCZ19,dabouei2020exploiting}, and adversarial training~\cite{MadryMSTV18,ZhangYJXGJ19,lee2020adversarial,deng2020adversarial}. 
Among which, adversarial training has been demonstrated as one of the most effective methods~\cite{AthalyeC018,LiLWZG19,dong2020benchmarking}. 
In the following, we focus on the discussion on adversarial training methods, and clarify how our method differs. 

%\textbf{Adversarial Training. }
Despite the great success of adversarial training (AT), there remains a visible robust generalization gap between training data and test data. 
Adversarial training with a specific attack would lead to weak adversarial robust generalization on test data~\cite{zhang2020attacks,deng2020adversarial}. 
Moreover, the adversarial robust generalization is theoretically more difficult than standard generalization~\cite{SchmidtSTTM18}, and often possesses significantly higher sample complexity~\cite{YinRB19,MontasserHS19}.% and needs more data. 

Recently, several AT-based methods have been proposed for improving the adversarial robustness.
\citet{ZhangYJXGJ19} propose TRADES to maximize the trade-off between adversarial robustness and standard accuracy, 
which is based on the locally most adversarial perturbed points and differs to ours.
\citet{lee2020adversarial} propose the adversarial vertex mixup by augmenting the adversarial examples through \textit{mixup} operation,
but they do not consider the diversity of adversarial directions as we do. 
They also adopt an existing label-smoothing  mechanism~\cite{szegedy2016rethinking} to regularize the classifier, while we design a distance-aware label smoothing (DLS) mechanism that is adaptive for treating the perturbed points differently based on their distances to the benign points. 

The most related method to ours is the adversarial distributional training (ADT)~\cite{deng2020adversarial}, which trains with a learned adversarial distribution through a neural network to characterize the potential adversarial examples for each benign sample. 
ADT treats the sampled perturbed training points equally, while we treat the sampled perturbed training points differently based on their characteristics.
In addition, modeling the adversarial distribution for each benign point in ADT is intrinsically hard and time-consuming, especially for the high-dimensional image data. 
While in our method, the sampling of perturbed training points is based on the commonly used PGD attack, 
therefore as compared with the standard adversarial training, almost no extra attack cost is introduced.

%\vspace{-0.5em}
\section{Methodology}
\label{sec:method}
We first present the background of the standard adversarial training (SAT)~\cite{MadryMSTV18}, then introduce in detail the proposed regional adversarial training (RAT).

\subsection{Standard Adversarial Training}
\label{sec:standard-at}

Given a standard classification task with training dataset $\mathcal{D} = \left\{(\bx_i, y_i)\right\}_{i=1}^{n}$ with $c$ classes and $n$ samples, where $\bx_i \in \mathbb{R}^d$ represents the benign point; $y_i \in \left\{1,\dots,c\right\}$ denotes the ground-truth class label, and let $\bm{y}_i$ be the one-hot form of the class label $y_i$. The standard adversarial training~\cite{MadryMSTV18} is formulated as a min-max optimization problem: %, which is expressed as:
\begin{equation}
	\label{eq: at_fomulation}
	\min_{\bm{\theta}} \ \frac{1}{n} \sum_{i=1}^{n} \max_{\bx'_i \in \mathcal{B}(\bx_i, \epsilon)} \mathcal{L}(f_{\bm{\theta}}(\bx'_i), \bm{y}_i),
\end{equation} 
where $f_{\bm{\theta}}(\cdot):\mathbb{R}^{d} \rightarrow \mathbb{R}^c$ is the classifier parameterized by $\bm{\theta}$ that outputs the predicted probabilities over all classes; $\mathcal{L}(\cdot,\cdot)$ is the cross-entropy loss; $\mathcal{B}(\bx, \epsilon)$ is the $\epsilon$-ball of the benign data point $\bx$, i.e., $\mathcal{B}(\bx, \epsilon) = \{\bx':\left \| \bx' - \bx \right \|_\infty \leq \epsilon \}$. Note that we mainly focus on the $\ell_{\infty}$ threat models to align with previous works.

As formulated in Eq.~\eqref{eq: at_fomulation}, the inner maximization of the SAT aims to generate the most adversarial perturbed point in  set $\mathcal{B}(\bx, \epsilon)$ for each benign point, and the outer minimization aims to minimize the cross-entropy loss of all the generated perturbed points.
Since the inner maximization does not have a closed-form solution for deep neural networks, a commonly used approximate method is the projected gradient descent (PGD) attack~\cite{MadryMSTV18}. Considering a benign sample $(\bm{x}, \bm{y})$, the PGD attack updates the adversarial perturbed points with multiple steps:
	\begin{equation}
		\bx'_{t+1} =\Pi_{\mathcal{B}(\bx, \epsilon)}\left(\bx'_{t}+\alpha \cdot \operatorname{sign}\left(\nabla_{\bx'_{t}} \mathcal{L}(f_{\theta}(\bx'_{t}), \bm{y})\right)\right),
	\end{equation}
where $\bx'_{t}$ is the perturbed point at the $t$-th step; $\Pi_{\mathcal{B}(\bx, \epsilon)}(\cdot)$ is the function to project the inputs into the allowed set $\mathcal{B}(\bx, \epsilon)$; $\alpha > 0$ is the step size of PGD attack. $\bx'_{0}$ is randomly sampled from $\mathcal{B}(\bx, \epsilon)$.

%We will introduce the proposed method based on the PGD attack below, but the extensions to other iterative attacks are also applicable.

By revisiting the formulation of SAT in Eq.~\eqref{eq: at_fomulation}, on one hand, we see that SAT always optimizes on the locally most adversarial perturbed point within the $\epsilon$-ball, overlooking the diversity of perturbed training points. 
Actually, for each benign data point, there usually exist multiple adversarial directions and magnitudes that could lead to successful attack.
On the other hand, for the outer minimization of SAT, all the perturbed training points are treated equally and assigned the same one-hot class label of their benign counterparts, while their characteristics are not utilized during the training.
	
\subsection{Regional Adversarial Training}

In this work, we explore the diversity of perturbed training points and treat them differently based on their characteristics. In general, our adversarial training framework could be expressed as: 
\begin{equation}
\begin{gathered}
\label{eq: at_fomulation_ours}
\min_{\bm{\theta}} \ \frac{1}{n \times m} \sum_{i=1}^{n} \sum_{j=1}^{m} \mathcal{L}(f_{\bm{\theta}}(\hat{\bx}_{i}^{j}), \hat{\bm{y}}_{i}^{j})), \\
\end{gathered}
\end{equation}
where
\begin{equation}
\begin{gathered}
\label{eq: at_fomulation_ours_label}
\hat{\bm{y}}_{i}^{j} = S_L(\hat{\bx}_{i}^{j}, \bm{y}_i).
\end{gathered}
\end{equation} 
Here $\hat{\bx}_{i}^{j}$ denotes the diverse perturbed training points for the benign sample $(\bm{x}_i, \bm{y}_i)$, where $j=1,...,m$; $S_{L}(\hat{\bx}_{i}^{j}, \bm{y}_i)$ is the \textit{soft label assignment function} for assigning representative soft labels to the perturbed training point $\hat{\bx}_{i}^{j}$ based on their characteristic.

To realize this framework, we propose a new defense method called \textit{Regional Adversarial Training (RAT)}. Firstly, in order to enrich the diversity of the perturbed training points, we propose an efficient \textit{Adversarial Region-based Sampler (ARS)}, where we construct an adversarial region around each benign data point based on the attack path of the PGD method, and sample perturbed training points with various directions and magnitudes inside this region. Secondly, 
%to take advantage of the diversity of these perturbed points, 
we capture the intuition that perturbed training points at different locations should have different contributions to the model performance. Thus we propose a \textit{Distance-aware Label Smoothing (DLS)} mechanism that assigns different smoothing factors to each perturbed training point based on their distance to the %corresponding 
benign point. 
In this way, our adversarially trained models make significant improvements on %the standard adversarial training
SAT, and exhibit better robust generalization.

In the following, we introduce in detail how ARS and DLS work, as well as the overall training procedure. 

\subsubsection{Adversarial Regional Sampler.}
%To consider the diversity of the perturbed training points during the adversarial training, we propose the adversarial regional sampler (ARS) to efficiently obtain diverse perturbed training points based on the PGD attack.
%We first define the {\it least adversarial point} $\bx_{least}'$ and the {\it most adversarial point} $\bx_{most}'$ of the PGD attack as follows.
To enrich the perturbed training points, one simple method is to generate a set of adversarial examples by performing PGD attack with various settings for multiple times. But it will significantly increase the training cost, especially for large-scale datasets. 
Thus, we propose an efficient adversarial region-based sampler (ARS) to obtain diverse perturbed training points.

To describe ARS more clearly, we define the {\it first adversarial point} $\bx_{f}'$ and the {\it end adversarial point} $\bx_{e}'$ of the PGD attack as follows.
%We define the {\it first adversarial point} $\bx_{f}'$ and the {\it last adversarial point} $\bx_{l}'$ of the PGD attack as follows.
%the first adversarial point and the last adversarial point along the attack path
\begin{definition} [First / End Adversarial Point]
Let $S(x) = \{\bm{x}_t'\}_{t=1}^K$ indicates the attack sequence of the benign sample $(\bm{x}, y)$ generated by a $K$-step PGD attack.
Assuming there exists at least one adversarial perturbed point that leads to a successful attack, 
%i.e., $\exists t \in \left\{1, \dots, K\right\}, s.t., \arg \min_{\hat{y} \in \mathcal{Y}}  \mathcal{L}(f_{\bm{\theta}}(\bx_{t}'),\hat{y}) \neq y$, 
the end adversarial point is defined as $\bx'_{e} \triangleq \bm{x}_K'$, and the first adversarial point is defined as $\bx'_{f} \triangleq \bm{x}_t'$, where 
%$t = \arg \min_{t, \bx_{t}' \in S(x)}  f_{\bm{\theta}}(\bx_{t}') \neq y$.
$t = \arg \min_{t}  (\phi(f_{\bm{\theta}}(\bx_{t}')) \neq y)$, and $\phi(\cdot)$ is the function to obtain the predicted label from output probability. 
%$j = \min t, s.t., \arg \min_{\hat{y} \in \mathcal{Y}}  \mathcal{L}(f_{\bm{\theta}}(\bx_{t}'),\hat{y}) \neq y$.
\end{definition}

In most cases, there will be at least one or two adversarial points along the path during the PGD process. Without loss of generality, we assume there exists at least one successful perturbed point along the PGD attack sequence.

%When the PGD attack could not achieve a successful attack point, the least and the most adversarial points would degenerate to the perturbed points $\bx_{K}'$ at the last PGD iteration. 
%While this degeneration situation is normal in robust training and has no effect to our training algorithm, we will demonstrate our ideas by assuming

As illustrated in Figure~\ref{fig: ball}, we construct a region with three points (the benign data point $\bx$, the first adversarial point $\bx'_f$ and the end adversarial point $\bx'_e$ along the PGD attack path). We observe that, firstly, starting from $\bx$ to any point on the line segment between $\bx'_f$ and $\bx'_e$, this region essentially provides diverse adversarial directions. Secondly, we could expand the region along the adversarial directions to construct a larger adversarial region for capturing richer perturbed points and sample multiple perturbed points with different magnitudes along each adversarial direction inside the adversarial region.

%The construction of ARS w.r.t. a benign sample $(\bm{x}, \bm{y})$ is illustrated in Figure~\ref{fig: ball}, where $\bar{\bx}$ denotes the point on the line segment between the least adversarial point $\bx'_{least}$ and the most adversarial point $\bx'_{most}$. 
%For the point $\bar{\bx}$, we have the following observations. \textbf{1)} The point $\bar{\bx}$ on the line segment generally possesses strong attack capacity, that is, points in the line segment between the least and the most adversarial points essentially provide diverse adversarial directions. \textbf{2)} For the adversarial direction from the benign point $\bx$ to the point $\bar{\bx}$, the attack capacity generally increases as the distance to the benign point $\bx$ increases. 

Based on the above observations, we propose the adversarial region-based sampler (ARS). 
ARS first randomly samples a point $\bar{\bx}$ on the line segment between $\bx'_{f}$ and $\bx'_{e}$ to determine the adversarial direction ($\bx$ to $\bar{\bx}$). Then, ARS samples a perturbed training point $\hat{\bx}$ with a magnitude scale factor $s$ along this direction. 
The sampling process is formulated as follows:
%As illustrated in Figure~\ref{fig: ball}, ARS first samples point $\bar{\bx}$ on the line segment between the first adversarial point $\bx'_{f}$ and the end adversarial point $\bx'_{e}$ to determine the adversarial direction, and then samples the perturbed training point $\hat{\bx}$ in the direction w.r.t. $\bar{\bx}$ with the adversarial magnitude determined by scale factor $s$: %, which is expressed as:
\begin{equation}
\begin{gathered}
\label{eq: sample}
\bar{\bx} = \lambda \cdot \bx'_{f} + (1 - \lambda) \cdot  \bx'_{e},\ \lambda \sim  \mathcal{U}(0,1), \\
    			\hat{\bx} =  \bx  + s  \cdot  (\bar{\bx} - \bx), \ s \sim \mathcal{S},
\end{gathered}
\end{equation}
where $\lambda$ is %sampled from $\mathcal{U}(0,1)$
uniformly sampled in [0,1]; $s$ is the scale factor to control the adversarial magnitude, i.e., the distance 
%between the perturbed training point $\hat{\bx}$ and the benign point $\bx$.
to the benign point $\bx$.
% The scale factor $s$  is sampled from the scale factor set $\mathcal{S}$\HK{where do you have the definition?}, and 
We define the magnitude scale factor from the benign sample $\bx$ to the line segment formed by $\bx'_{f}$ and $\bx'_{e}$ is $1$. The magnitude scale factor $s$ is randomly 
picked from the scale candidate set $\mathcal{S}$, 
%sampled in $[0,S]$ %\HK{I feel in Eq. (5), $s \sim \mathcal{S}$ need to be changed to $s \sim \mathcal{U}(0,S)$ }
which consists of multiple scale factors varying from $0$ to $S (S = 2$ or 3 in our experiments). A larger scale factor $s$ indicates that the sampled perturbed training point $\hat{\bx}$ is farther away from the corresponding benign point $\bx$. 
When $s$ is 0, the sampled perturbed point $\hat{\bx}$ coincides with the benign point $\bx$. 
When $s$ is 1, $\hat{\bx}$ lies in the farthest bound of the expanded adversarial region. 

In this way, for each training iteration of RAT, ARS only needs to perform the PGD attack once to generate the perturbed training points with diverse directions and magnitudes, and no extra attack cost is imposed as compared to the standard adversarial training. 

\begin{figure}[t]
\begin{center}    			
\includegraphics[width=0.4\textwidth]{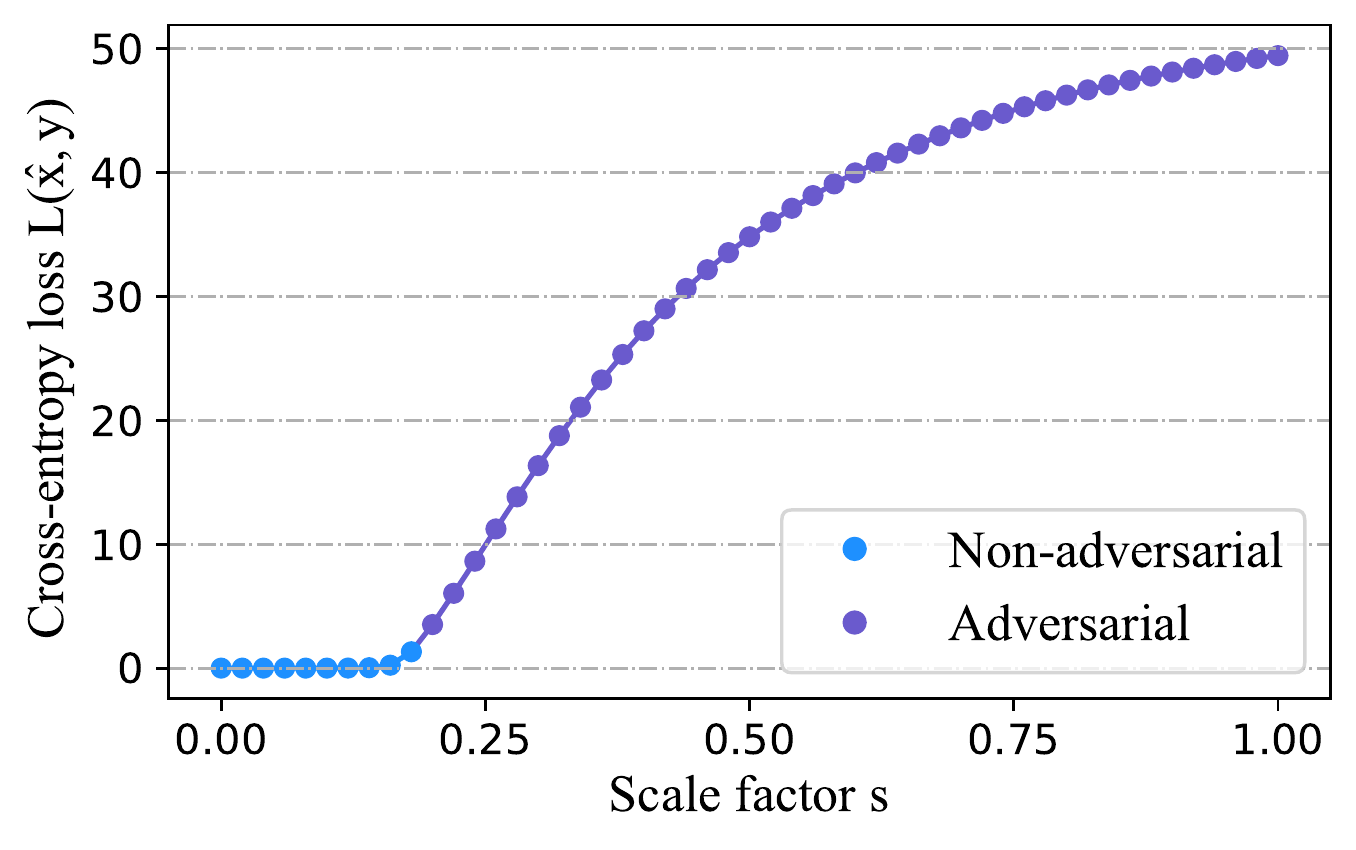}
\end{center}
\vspace{-0.2em}
\caption{Visualization of the cross-entropy loss and attack result for the perturbed point $\hat{\bx}$ sampled in a selected adversarial direction with various magnitude scale factor $s$. Here \textit{Adversarial} denotes successful attack, and \textit{Non-adversarial} denotes unsuccessful attack.}
\label{fig: obervation}
\vspace{-0.5em}	
\end{figure}

\subsubsection{Distance-aware Label Smoothing.}

Directly adopting the one-hot class label of the benign samples to train with diverse perturbed points may overlook their characteristics and differences, leading to the underutilization of the diversity.
Our intuition is that %perturbed training points at different locations should have different contributions to the model performance, with the principle that
a perturbed point closer to the benign point should have higher confidence w.r.t. the ground-truth class label, and vise versa.

To verify our intuition, we randomly pick a correctly classified sample $(\bm{x}, \bm{y})$ on CIFAR-10, and then sample the perturbed point $\hat{\bx}$ with multiple magnitudes along the randomly selected direction. 
%the direction determined by a randomly sampled $\bar{\bx}$ with the scale factor $s$, i.e., $\hat{\bx} =  \bx  + s  \cdot  (\bar{\bx} - \bx)$, and visualize the cross-entropy loss and attack results of the perturbed point $\hat{\bx}$.
We visualize the cross-entropy loss and attack results of the perturbed point $\hat{\bx}$ in Figure~\ref{fig: obervation}.
It shows that the cross-entropy loss increases as the magnitude scale factor $s$ increases, which demonstrates our intuition.
%that we can sample the perturbed points with diverse adversarial magnitudes in the direction w.r.t. point $\bar{\bx}$, where $\bar{\bx}$ determines the sampled adversarial direction, and the scale factor $s$ determines the sampled adversarial magnitude.

Based on the above observations, we propose the distance-aware label smoothing (DLS) mechanism to treat the perturbed training points sampled from ARS differently.
%, with the principle that a perturbed point $\hat{\bx}$ closer to the benign point should have higher confidence w.r.t. the ground-truth class label, and vise versa.
%Specifically, 
Considering a perturbed training point $\hat{\bx}$ sampled based on Eq.~\eqref{eq: sample}, the corresponding distance-aware soft label $\hat{\bm{y}}$ is formally defined as follows:
\begin{equation}
\begin{gathered}
\label{eq: label_y}
\hat{\bm{y}} = S_{L}(\hat{\bx}, \bm{y}) = \bm{y} \cdot  \beta + (1 - \bm{y} ) \cdot \frac{1 - \beta}{c-1} ,
\end{gathered}
\end{equation}
where
\begin{equation}
\begin{gathered}
\label{eq: label_beta}
\beta = \beta_{max} - s  \times \frac{\beta_{max} - \beta_{min}}{S}. %{|\mathcal{S}|_{\infty}}.
\end{gathered}
\end{equation}
$\bm{y}$ is the one-hot ground-truth class label of the benign point $\bm{x}$; $c$ is 
the number of classes, and
$s$ is the magnitude scale factor for the perturbed point $\hat{\bx}$ in Eq.~\eqref{eq: sample}; 
%$|\mathcal{S}|_{\infty}$ 
$S$ is the maximal magnitude scale factor inside the adversarial region; %value among the set $\mathcal{S}$;
$\beta$ is the label smoothing factor determined by the scale factor $s$;
$\beta_{max}$ and $\beta_{min}$ are the maximal and minimal label-smoothing factors, respectively. As can be seen in Eq.~\eqref{eq: label_beta}, a larger scale factor $s$ can lead to a smaller label smoothing factor $\beta$, and a lower confidence w.r.t. the ground-truth class label $y$ for the perturbed training point $\hat{\bx}$.
As shown in Figure~\ref{fig: ball}, compared to $\hat{\bx}^1$, the sampled perturbed point $\hat{\bx}^2$ farther away from the benign point $\bx$ has a lower confidence w.r.t the ground-truth class label.

\subsubsection{Overall Training Procedure.}
The overall training procedure of %the proposed regional adversarial training (RAT)
RAT 
is summarized in Algorithm~\ref{alg: reg}. %and the generation procedure of the perturbed training samples for RAT is summarized in Algorithm~\ref{alg: ars}.
During the training, RAT first obtains the attack path for each benign point based on the PGD attack. Then, RAT samples multiple perturbed training points with random adversarial directions and magnitudes based on ARS. Third, RAT assigns such perturbed training points with the distance-aware soft labels based on DLS. In the end, RAT computes the loss of the perturbed training samples and updates the model parameters $\bm{\theta}$ with the gradient on loss. 

Overall, the framework of RAT facilitates a more reasonable learning scheme for adversarial training by considering the diversity and characteristics of the perturbed training points. 
Note that the standard adversarial training is a special case of RAT, by specifying the $\bx'_f \triangleq \bx'_e$ for the adversarial region, the scale factor $s=1$ for the adversarial regional sampler, and the label smoothing factors $\beta_{min} = \beta_{max} =1.0$ for the distance-aware label smoothing mechanism.

\begin{algorithm}[t]
	\caption{Regional Adversarial Training (RAT)}
	\label{alg: reg}
	\begin{algorithmic}[1]
		\REQUIRE  Network $f_{\bm{\theta}}$, training dataset $\mathcal{D}$, training epochs $T$,  number of batches $M$, batchsize $n$, learning rate $\eta$ and the number of samples $m$
		\FOR{epoch = $1,...,T$}
		\FOR{mini-batch = $1,...,M$}
		\STATE Sample a mini-batch $\{(\bx_i, \bm{y}_i)\}^n_{i=1} \sim \mathcal{D}$
		\FOR{$i=1$ to $m$ (in parallel)}
		\STATE Generate multiple perturbed training points and their distance-aware soft labels $\{(\hat{\bx}_i^j,\hat{\bm{y}}_i^j)\}_{j=1}^{m}$ for $(\bx_i, \bm{y}_i)$ by Eq.~\eqref{eq: sample} and Eq.~\eqref{eq: label_y}
		\ENDFOR
		\STATE $\bm{\theta} \leftarrow \bm{\theta} - \eta \cdot \frac{1}{n \times m}\sum_{i=1}^{n}\sum_{j=1}^{m} \nabla_{\bm{\theta}} \mathcal{L}(f_{\bm{\theta}}(\hat{\bx}^j_{i}), \hat{\bm{y}}^j_{i}) $
		\ENDFOR
		\ENDFOR
		\ENSURE robust network  $f_{\bm{\theta}}$
	\end{algorithmic}
\end{algorithm}

\vspace{-0.5em}
\section{Experiments}
\label{sec: exp}
To evaluate the robustness and generalization ability of the proposed RAT, we present experimental results on various datasets and models. All image values are normalized into $[0, 1]$. All experiments are run on a single NVIDIA Tesla V100 GPU. 

\textbf{Baseline Methods.} 
We compare RAT with the following training methods: standard training (ST), standard adversarial training (SAT)~\cite{MadryMSTV18}, TRADES~\cite{ZhangYJXGJ19}, adversarial vertex mixup (AVM)~\cite{lee2020adversarial}, and adversarial distributional training (ADT)~\cite{deng2020adversarial}. We adopt  $\lambda$ = 6 for TRADES, and adopt ADT\textsubscript{EXP} version for ADT as reported in the original paper for the best performance.

\textbf{Datasets.}
We use three benchmark datasets including CIFAR-10~\cite{krizhevsky2009learning}, CIFAR-100~\cite{krizhevsky2009learning}, and a subset of ImageNet called ImageNette~\cite{imagenette}. CIFAR-10 and CIFAR-100 are  widely adopted by adversarial training studies including the baseline methods. For the large-scale ImageNet dataset, just as all the baseline methods did not report the results, we are also unable to evaluate on ImageNet due to the very high training cost. Still, we add a comparison on ImageNette, a subset of ImageNet.

\textbf{Training Setup.} 
For the neural networks, we choose WRN-34-10~\cite{zagoruyko2016wide} for the experiments on CIFAR, and PreActResNet18~\cite{he2016identity} for the experiments on ImageNette.
For fair comparisons, we follow the training settings as in~\cite{MadryMSTV18, lee2020adversarial}, and train the models using SGD with 0.9 momentum for $80k$ steps with the initial learning rate of
0.1, but divided by 10 at the step $30k$ and $45k$, respectively. The batch size is 128 and the weight decay is $2 \times 10 ^{-4}$. For the generation of adversarial examples, the perturbation bound $\epsilon$ = 8/255; the PGD iteration number $K$ = 10; and the PGD step size $\alpha$ = 2/255. 
For the hyperparameters of RAT, we set the scale factor set $\mathcal{S}$ as %$\left[0.0 : 0.1 : 2.0\right]$
$\{0.0, 0.1, 0.2, ..., 2.0\}$
for CIFAR and %$\left[0.0 : 0.1 : 3.0\right]$
$\{0.0, 0.1, 0.2, ..., 3.0 \}$
for ImageNette, the number of samples $m$ = 2, and the maximal and minimal label-smoothing factors are set to 1.0 and 0.1, respectively. 

\begin{table}[t]
		\centering
		\footnotesize
		\vspace{5pt}
		\setlength{\tabcolsep}{2.5mm}{
		\scalebox{0.95}{	\begin{tabular}{l|ccccc} 
				\toprule
				Model & Clean & FGSM & PGD & CW$_\infty$ & $\mathcal{N}$attack \\
				\midrule
	       	\multicolumn{6}{c}{\textbf{CIFAR-10} (The perturbation bound $\epsilon = 8/255$).} \\
				\midrule
				ST & \textbf{95.42} & 12.91 & 0.00 & 0.00 & 0.00\\
				SAT & 86.92 & 55.58 &47.74 & 48.64 & 47.20 \\
				TRADES & 85.84 & 59.23 & 50.39 & 51.02 & 50.30\\
				AVM & 92.40 & 76.92 & 62.19 & 56.83 & 58.00 \\
				ADT & 87.98 & 58.25 & 49.27 & 50.38 & 47.60 \\
				RAT & 93.17 & \textbf{77.51} & \textbf{70.57} & \textbf{65.54} & \textbf{69.70}\\
				\midrule
				\multicolumn{6}{c}{\textbf{CIFAR-100} (The perturbation bound $\epsilon = 8/255$).} \\
				\midrule
				ST & \textbf{78.75} &6.99 & 0.00 & 0.00 &  0.00 \\
				SAT & 60.03 & 28.27 & 24.34 & 23.95 & 26.50 \\
				TRADES & 57.64 & 29.80 & 26.43 & 25.25 &  27.40\\
				AVM &  69.48 & 52.95 & 29.53  & 24.00 &  30.90 \\
				ADT & 62.63 & 29.94 & 24.53 &  25.04 &  27.10\\
				RAT & 70.65 & \textbf{57.16}  & \textbf{37.03} & \textbf{27.45} & \textbf{40.70} \\
				\midrule
				\multicolumn{6}{c}{\textbf{ImageNette} (The perturbation bound $\epsilon = 8/255$).} \\
				\midrule
				ST & \textbf{89.02} & 7.80 & 0.00 & 0.00 &  0.00\\
				SAT & 83.52 & 57.76 & 53.58 & 52.92 & 50.90\\
				TRADES & 82.52 & 58.01	& 53.73 & 53.25 & 52.20\\
				AVM & 87.87 & 58.55 & 50.14 & 45.22 &  54.10 \\
				ADT & 78.83 & 56.05 & 52.99 & 51.82 & 51.50 \\
				RAT & 86.19 & \textbf{74.75} & \textbf{61.83} & \textbf{55.82} & \textbf{59.60} \\
				\bottomrule
		\end{tabular}}}
		\caption{Test accuracy (\%) of RAT and the baselines under the white-box and black-box attacks on CIFAR-10, CIFAR-100 and ImageNette. Clean denotes the standard accuracy.}
		\label{tab: all_results}
\vspace{-0.5em}		
	\end{table}

\begin{figure}[t!]
\begin{center}
\includegraphics[width=0.4\textwidth]{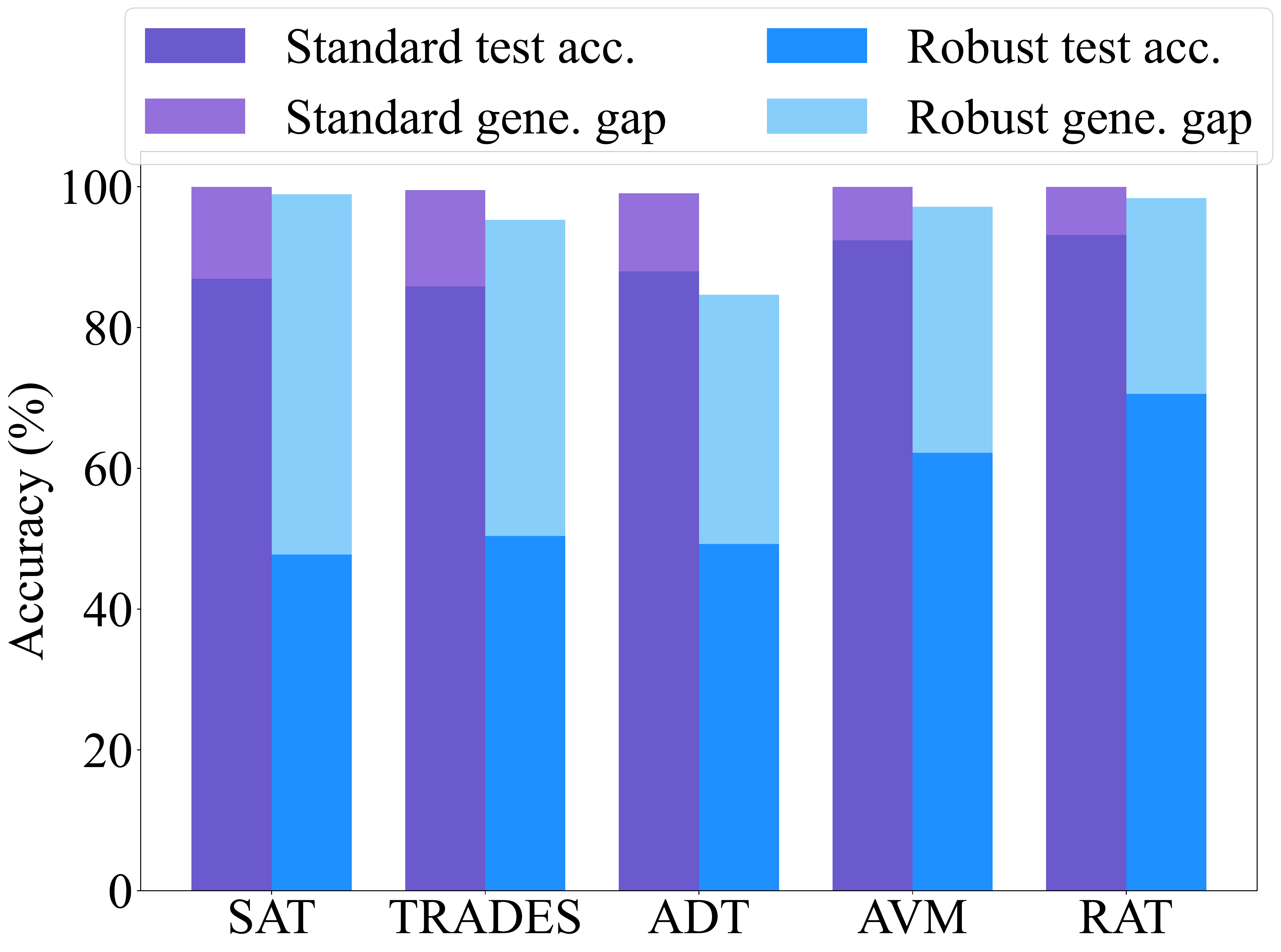}%width=0.42
\end{center}
\caption{Standard and adversarially robust generalization gaps of RAT and 
%the competing defense methods 
the baselines 
on CIFAR-10.}
\label{fig: genecifar}
\vspace{-1em}	
\end{figure}

\textbf{Evaluation Setup.}
We report both the standard accuracy on clean data, and the robust accuracy on adversarial examples. Following the widely adopted protocol~\cite{MadryMSTV18, ZhangYJXGJ19}, we consider the white-box attacks including FGSM~\cite{GoodfellowSS14}, PGD~\cite{MadryMSTV18} and CW$_\infty$~\cite{carlini2017towards} %and FAB~\cite{Croce020}
, and the black-box attack, $\mathcal{N}$attack~\cite{LiLWZG19}. For PGD and CW$_\infty$, the perturbation bound $\epsilon$ = 8/255; the step number $K$ = 10; and the step size $\alpha$ = 2/255, which keep the same as in the training setting. For the $\mathcal{N}$attack, we perform on a subset of 1,000 randomly selected test images due to the high complexity for query, and set the maximum iteration number to 200 and sample 300 perturbations at each iteration.

\subsection{Evaluation Results}
We first validate our main intuitions: by considering the diversity and characteristics of the perturbed training points, is it possible to develop the robust models that show better robust generalization?

As shown in Table~\ref{tab: all_results}, we report the standard and robust test accuracy of RAT and the baselines on three standard datasets. Clearly, RAT achieves the best adversarial robustness against both white-box and black-box attacks, and maintains high standard test accuracy on various datasets. For example, on CIFAR-10 dataset, RAT outperforms the robust accuracy of the SAT baseline by a large margin of 22.83\% under the PGD attack,
meanwhile RAT outperforms the standard accuracy of SAT by a clear margin of 6.25\%. 
These results indicate that, by considering the diversity and characteristics of the perturbed training points, RAT could significantly improve the robust generalization of SAT, verifying that the framework behind RAT is more suitable for adversarial training.

%\vspace{-0.8em}
\subsection{Further Analysis on RAT}
%Besides the robustness evaluation under the white-box and black-box attacks, we also conduct a series of additional analysis to further analysis the robust generalization of RAT.
We conduct a series of additional analysis on the robust generalization of RAT.

%\textbf{Effect on Generalization Gaps.} 
\textbf{Comparison on Generalization Gaps.} We first analyze the standard and adversarial robust generalization gaps for RAT and the baseline defense methods. The standard generalization gap is defined as the difference of the standard accuracy between training data and test data, and similarly, the adversarial robust generalization gap is defined as the difference of the robust accuracy on the PGD attack between training data and test data. 
The results on CIFAR-10 dataset are reported in Figure~\ref{fig: genecifar}. 
We have similar results on CIFAR-100 and ImageNette, as reported in the supplementary material. 
These results consistently demonstrate that RAT can bring smaller gaps on the standard and adversarial robust generalization, as compared to other AT-based methods.

\begin{figure}[tbp]
\begin{center}
\includegraphics[width=0.4\textwidth]{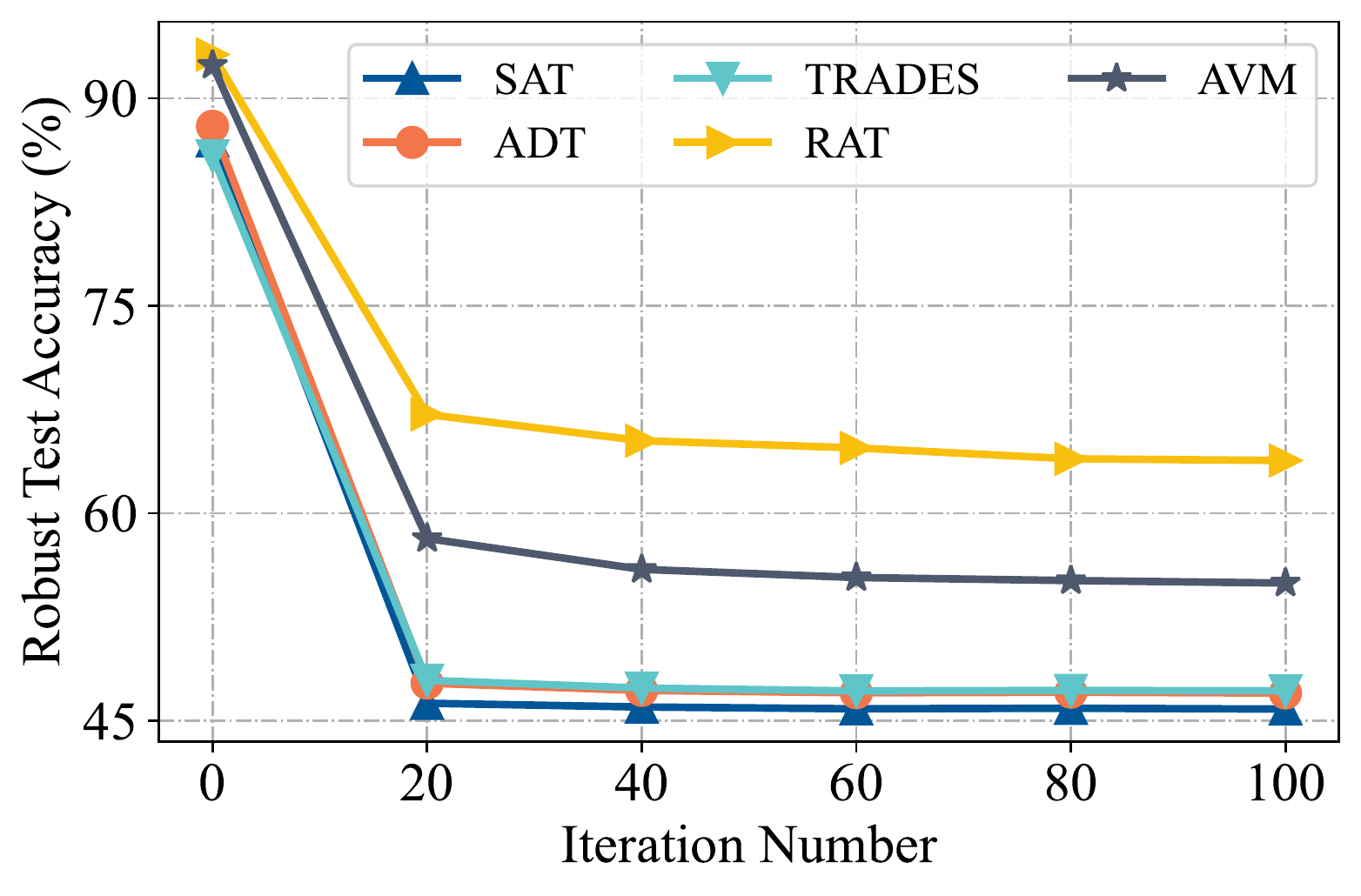} %width=0.445
\end{center}
\caption{Robust test accuracy (\%) of  RAT and the baselines against PGD with various iterations on CIFAR-10. The PGD perturbation budget is 8/255 and the PGD step size is 2/255.}
\label{fig: cifar10_varying_step}
\vspace{-0.5em}	
\end{figure}

\begin{figure}[tbp]
\begin{center}
\includegraphics[width=0.4\textwidth]{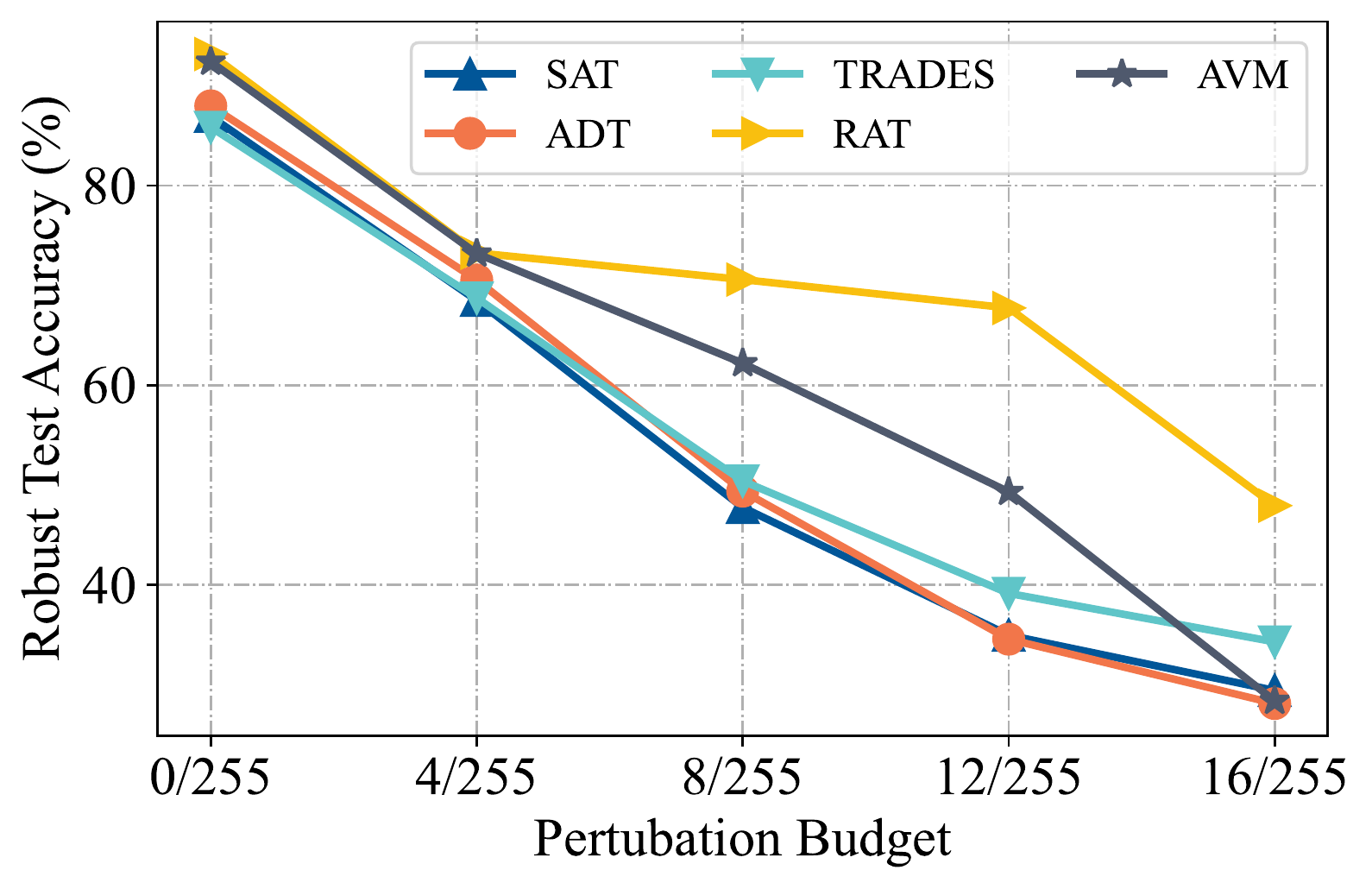} %width=0.445
\end{center}
\caption{Robust test accuracy (\%) of RAT and the baselines against PGD with various perturbation budgets on CIFAR-10.
The PGD step size is 2/255, the  iteration number is 10.}
\label{fig: cifar10_varying_eps}
\vspace{-0.5em}	
\end{figure}

\textbf{Robustness against PGD with Various Iterations and Perturbation Budgets.} 
%We %further analyze the robustness against PGD with various number of attack iterations $K$ for RAT and the baselines. 
1) As shown in Figure~\ref{fig: cifar10_varying_step}, we visualize the robust accuracy of RAT and the baselines under the PGD attack with the iteration number $K$ varying from 0 to 100, where the perturbation budget is 8/255 and the step size is 2/255. %We can see that, 
The robust accuracy of all the defense methods first decreases and then stabilizes as the iteration number increases. 
RAT consistently outperforms all the baselines under various iteration numbers, and outperforms the second best by a clear margin of around 10\% after 20 steps of PGD iterations.
2) As shown in Figure~\ref{fig: cifar10_varying_eps}, we visualize the robust accuracy against PGD with the perturbation budget varying from 0 to 16/255, where the step size is 2/255 and the iteration number is 10. 
%It is obvious that 
The robust accuracy of all the defense methods decreases along with the increasing perturbation budget. 
However, RAT exhibits stronger adversarial robustness, and the difference is much clearer for higher perturbation budgets. 
We have similar results on CIFAR-100 and ImageNette, as reported in the supplementary material. 
These results consistently indicate that RAT can achieve better robust generalization on the PGD attack with different iteration numbers and perturbation budgets. 
%even though it only adopts the PGD attack with $10$ iteration steps during the training. 

% \textbf{Robustness against PGD with various Perturbation Budgets.}
% %Besides evaluating on the PGD attack with various iterations, 
% We further analyze the model robustness under the PGD attack with various perturbation budgets $\epsilon$. 
% As shown in Figure~\ref{fig: cifar10_varying_eps}, we visualize the robust accuracy against PGD with the perturbation budget varying from 0 to 16/255, where the step size is 2/255 and the iteration number is 10. 
% It is obvious that the robust accuracy of all the defense methods decreases along with the increasing perturbation budget. 
% However, RAT exhibits stronger adversarial robustness, and the difference is much more clear for higher perturbation budgets. 
% %For instance, RAT outperforms the robust accuracy of SAT by 32.79\% under the PGD attack with the perturbation budget 12/255. 
% We have similar results on CIFAR-100 and ImageNette, as reported in the supplementary material.
% These results consistently verify that RAT can achieve better adversarial robust generalization on the typical PGD attack with various perturbation budgets. 
% %For the reason, We suspect it is that RAT adopts a large scale factor $s$ across various adversarial directions, and trains with stronger perturbed training samples outside the $\epsilon$-ball of the benign points.

\begin{table*}[t]
		\footnotesize
		\vspace{-4pt}
		\begin{center}
			\setlength{\tabcolsep}{1.7mm}{
					\scalebox{0.95}{	\begin{tabular*}{\textwidth}{ @{\extracolsep{\fill}} lccccccccccccc}
					\toprule
					\multirow{2}[4]{*}{Model} & \multirow{2}[4]{*}{mCA} &\multicolumn{4}{c}{Blur Corruptions}  & \multicolumn{4}{c}{Weather Corruptions} & \multicolumn{4}{c}{Digital Corruptions} \\
					\cmidrule(lr){3-6} \cmidrule(lr){7-10} \cmidrule(lr){11-14} 
					& & Defocus &  Glass  &  Motion  &  Zoom  & Snow  &  Frost  &  Fog   & Bright &Contrast &Elastic& Pixelate &JPEG  \\
					\midrule
			   	 \multicolumn{14}{c}{\textbf{CIFAR-10-C}} \\
					\midrule
					ST & 81.15 & 83.40 & 53.27 & 79.51 & 77.23 & 83.57 & 82.03 & \textbf{90.59} & \textbf{94.42} & \textbf{92.68} & 84.18 & 75.72 & 77.19\\
					SAT & 81.58 & 83.05 & 79.97 & 80.02 &82.20 & 82.40 &  85.26  & 71.70  & 86.70 & 76.38 & 81.13 &  85.07 & 85.06 \\
					TRADES & 80.67 & 82.34 &  79.02 & 79.32 & 81.41 & 81.00 & 84.22 & 71.13 & 85.83 & 75.80 & 80.24 &  83.94 & 83.80 \\
					AVM & 87.05 & 88.52 &  82.67 & 85.24 & 87.57 & 86.98 & 89.50 & 81.29 &  91.85 & 84.12 & 86.91 & 89.99 & 89.96 \\
					ADT & 77.36 & 82.76 & 80.30 & 78.32 & 81.59 & 81.49& 78.51 & 62.54 & 84.52 & 45.67 & 81.50 & 85.66 & 85.51  \\
					RAT & \textbf{87.94} &  \textbf{89.12} & \textbf{83.25} & \textbf{85.75} &  \textbf{88.45} & \textbf{87.90} & \textbf{90.15} & 83.04 & 92.76 &  85.70 & \textbf{87.95} & \textbf{90.35} & \textbf{90.86}  \\
					\midrule
			   	 \multicolumn{14}{c}{\textbf{CIFAR-100-C}} \\
					\midrule
					ST & 57.45& 62.73 & 21.30&58.27 & 57.47& 56.63& 52.75&\textbf{69.95} &\textbf{75.46} & \textbf{71.89}& 61.79&  54.13& 46.98 \\
					SAT &52.87 & 54.84 &50.85 & 51.30&53.13 &52.80 &56.76 &41.80 &  59.07&47.26 &52.30 & 57.59& 56.71 \\
					TRADES &50.99 & 52.89&50.15 & 49.57& 51.60 &  51.04 &  54.81& 39.64&  57.26&44.85 &50.26 &55.35 &54.47 \\
					AVM &61.48 & 63.77 &  55.37 &59.51 & 62.39& 61.26& 64.46& 52.62&  67.75& 57.68&  61.38 &65.75 & 65.86\\
					ADT &50.79 &55.97 &52.69 &51.56 & 54.75 & 53.37& 49.66& 34.80& 57.99& 25.37& 54.30&60.02 & 59.04 \\
					RAT & \textbf{62.44} & \textbf{64.20} &  \textbf{55.39} & \textbf{59.84} & \textbf{62.70} & \textbf{62.20} & \textbf{65.71}& 54.55 & 69.27& 59.92 &\textbf{62.00} &\textbf{67.30} & \textbf{66.21} \\
					\bottomrule
			\end{tabular*}}}
		\caption{Classification accuracy (\%) of RAT and the baselines on the commonly occurring natural image corruptions of CIFAR-10-C and CIFAR-100-C. The \textit{mCA} refers to the mean accuracy averaged on different corruptions and severities.}
		\label{tab: cifar_10_c_result}
		\end{center}
    \vspace{-1.5em}
	\end{table*}

\begin{table}[h]
\centering
\footnotesize
\vspace{4pt}
\setlength{\tabcolsep}{3.8mm}{
\scalebox{0.95}{
\begin{tabular}{l|ccc} 
\toprule
Model  & Clean & PGD & CW$_\infty$ \\
\midrule
SAT  &  86.92 &  47.74 & 48.64 \\
SAT+ARS & 87.67 & 52.59 & 50.87  \\ 
SAT+ARS+DLS & \textbf{93.17} & \textbf{70.57} & \textbf{65.34}  \\ 
\bottomrule
\end{tabular}}}
\caption{Ablation study for the impact of ARS and DLS.}
%the adversarial regional sampler (ARS) and the distance-aware label smoothing (DLS).}
\label{tab: ablation}
\vspace{-0.8em}
\end{table}
	
\vspace{-0.1em}
\begin{table}[h]
		\centering
		\footnotesize
		\vspace{4pt}
		\setlength{\tabcolsep}{2.6mm}{
		\scalebox{0.95}{	\begin{tabular}{l|cc|ccc} 
				\toprule
				Method & $\beta_{max}$  & $\beta_{min}$ & Clean & PGD & CW$_\infty$  \\
				\midrule
			\multirow{5}{*}{RAT}	& 1.0 & 0.1 &  90.94&  \textbf{70.57} &   \textbf{65.54}  \\
			  & 1.0  & 0.3 & \textbf{92.59} &  64.87 &  61.53 \\
			 &1.0 & 0.5 &  91.55 & 62.86 &  56.95 \\
			  & 0.8 & 0.1 &92.33  &   66.44 & 63.47 \\
			 & 0.6 & 0.1 &91.65 & 62.93 & 59.88 \\
				\bottomrule
		\end{tabular}}}
		\caption{Test accuracy (\%)  of the training framework RAT with different maximal label-smoothing factor $\beta_{max}$ and minimal label-smoothing factor $\beta_{min}$ on  CIFAR-10.}% the CIFAR-10 dataset.}
		\label{tab: ab_beta}
\vspace{-0.8em}		
\end{table}

\textbf{Robustness against Natural Image Corruptions.}
Recent works~\cite{GilmerFCC19,YinLSCG19} found that adversarial training may harm the robustness against natural image corruptions, as compared with standard training. Therefore, to further investigate the performance of RAT, we consider to quantify the robustness under the natural image corruptions. Specifically, we consider the benchmark datasets CIFAR-10-C~\cite{HendrycksD19} and CIFAR-100-C~\cite{HendrycksD19} that consists of various types of natural image corruptions. 
We select 12 types of common image corruptions, including blur corruptions, weather corruptions and digital corruptions. For each type of corruptions, there are five levels of severity. 

As shown in Table~\ref{tab: cifar_10_c_result}, we report the average accuracy of all the five levels of severity for CIFAR-10-C and CIFAR-100-C, respectively. We can easily see that RAT leads to better robustness under a wide range of natural image corruptions, and outperforms all the baselines in terms of the average accuracy over all corruptions. The evaluations indicate that RAT can prevent the models from overfitting to certain attacking patterns, and exhibits better robust generalization on natural image corruptions.

\textbf{Identifying Obfuscated Gradients.} 
Here we further show that RAT does not lead to the obfuscated gradients~\cite{AthalyeC018, LiLWZG19} as RAT does not fit the following characteristics defined by ~\citet{AthalyeC018} and ~\citet{abs-1807-06732} that models with obfuscated gradients may suffer. 

%Recently, Athalye et al.~\cite{AthalyeC018} and Li et al.~\cite{LiLWZG19} found a phenomenon of \textit{obfuscated gradients} that some defense methods provide few benefits for the true robustness but mask the gradients on which most attacks rely. 
%Such phenomenon is named as \textit{obfuscated gradients}. %Below, we discuss how our defense mechanism not lead to obfuscated gradients on the basis of characteristics defined in ~\cite{AthalyeC018,abs-1807-06732}. 
%We argue that our defense mechanism does not lead to obfuscated gradients because it does not fit the characteristics defined by ~\citet{AthalyeC018} and ~\citet{abs-1807-06732}. 

1) \textit{One-step attacks perform better that iterative attacks perform.} On the contrary, extensive evaluations in Table~\ref{tab: all_results} indicate that under the defense of RAT, stronger iterative attacks (e.g., PGD and CW$_\infty$) are more powerful than the single-step attack (e.g., FGSM).
2) \textit{Robustness under white-box setting is higher than that under black-box setting.} 
%Since the adversary in the white-box setting has complete knowledge of the model, if the model does not suffer from the obfuscated gradients, the model robustness under white-box setting should be inferior to that under black-box setting. 
The results of RAT in Table~\ref{tab: all_results} and Figure~\ref{fig: cifar10_varying_step} show that, the stronger iterative white-box attacks are more successful than black-box attacks, which contradicts the model's gradient being obfuscating.   
3) \textit{A higher distortion bound increases the defense robustness.} %The success rate of attack should increase monotonically with the increase of the perturbation budget. 
Conversely, the evaluation in Figure~\ref{fig: cifar10_varying_eps} shows that the robustness of RAT decreases with the increase of the perturbation budget.

\textbf{Computation Efficiency.} The time cost of adversarial training is mainly caused by the attack overhead. RAT only needs to execute one PGD attack in each training iteration, thus RAT does not introduce extra attack cost as compared with SAT. For instance, on CIFAR-10, the running time of 1000 training iterations is $41.88$ minutes for SAT and $47.95$ minutes for RAT, 
%It can be seen that
thus little training cost is introduced for each training iteration of RAT.

%In summary, RAT follows the defined characteristics and therefore does not obfuscate gradients.

\subsection{Ablation Studies}
To gain further insights on the performance of RAT, we delve into RAT to investigate the impact of various components (i.e., the adversarial regional sampler (ARS) and the distance-aware label smoothing (DLS)) on CIFAR-10. We report the standard accuracy and the robust accuracy over the PGD, CW${_\infty}$ attacks for each model.

\textbf{Impact of ARS and DLS.} Since DLS is dependent with ARS in RAT, we train the models of SAT, the combination of SAT+ARS, and the combination of SAT+ARS+DLS (i.e., RAT) following the same training setting in Section~\ref{sec: exp}, to show the impact of ARS and DLS. 
As shown in Table~\ref{tab: ablation}, we can see that SAT+ARS exhibits great improvements on SAT over both the standard accuracy and the robust accuracy, which indicates considering the diversity of the perturbed training points does help promote the adversarial robustness without the degradation of standard accuracy. After further combining SAT+ARS with DLS, we can see that DLS further takes the advantage of  ARS, and yields more improvements on SAT, which demonstrates that treating the perturbed training points based on their characteristics can achieve better robust generalization.

\textbf{Selection of Maximal and Minimal Label-smoothing Factors.} 
Different selections of the maximal and minimal label-smoothing factors ($\beta_{max}$ and $\beta_{min}$) in RAT lead to different trade-offs between the standard accuracy and the adversarial robustness of the trained models, as shown in Table~\ref{tab: ab_beta}. 
Such trade-off is prevalent w.r.t. the hyperparameter settings in various adversarial training frameworks~\cite{MadryMSTV18, ZhangYJXGJ19, deng2020adversarial}. Besides, RAT makes steady improvements for different maximal and minimal label-smoothing factors in a wide range.

\section{Conclusion}
To boost the adversarially robust generalization, we explore a new adversarial training framework that considers the diversity as well as characteristics of the perturbed points in the neighborhood of benign samples, and implement our idea by proposing 
%a novel defense method called the Regional Adversarial Training (RAT).
a Regional Adversarial Training (RAT) defense method. 
The key components of RAT are the adversarial region-based sampler (ARS) that aims to consider the sampling diversity, and the distance-aware label smoothing (DLS) that aims to treat the sampled perturbed training points differently based on their characteristics.
Extensive experiments show that RAT could efficiently enhance the adversarial robustness without extra training cost as compared with the standard adversarial training, and performs favorably against state-of-the-art defenses in terms of standard accuracy and adversarial robustness. %, suggesting the significance of the diversity of the perturbed training points and the utilization of their characteristics for reducing the adversarial robust generalization. 
RAT also performs more stably with better robustness on the PGD attack with various settings and natural image corruptions, as compared to other AT-based methods.
%RAT offers a way to realize and validate the proposed new adversarial training framework, and we will investigate other possible realizations in future work.  

%In future work, we will investigate other possible realizations based on the new adversarial training framework.

% Use \bibliography{yourbibfile} instead or the References section will not appear in your paper
\bibliography{aaai22}

\newpage
\section*{Appendix}
% \newpage

% \section*{Appendix}
We report more experimental results on the CIFAR-100 and ImageNette datasets below, including the comparison on generalization gaps, and the robustness against PGD with various iterations and perturbation budgets.

\begin{figure*}[ht]
\begin{center}
\includegraphics[width=0.90\textwidth]{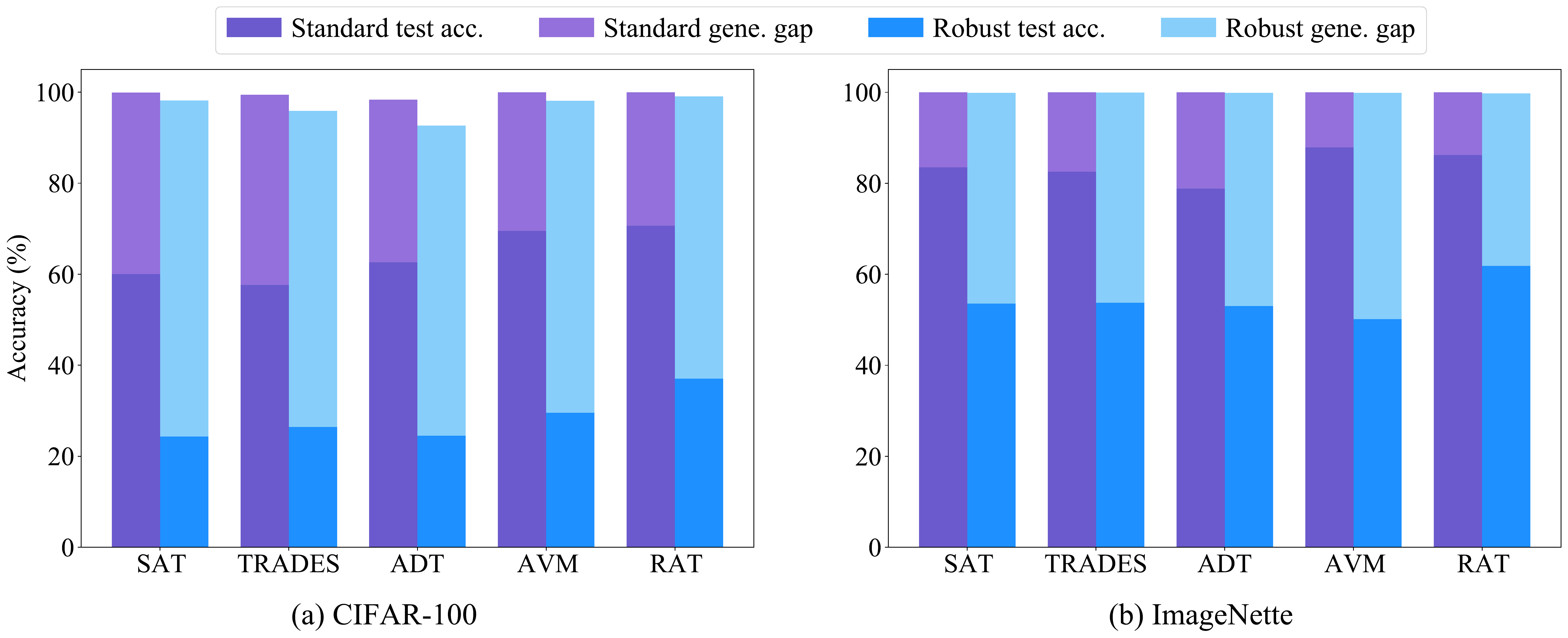}%width=0.42
\end{center}
\vspace{-0.8em}
\caption{Standard and adversarially robust generalization gaps of RAT and the baselines on CIFAR-100 and ImageNette.}
\label{fig: gene_other}
\end{figure*}

\smallskip
\textbf{Comparison on Generalization Gaps.} Similar to the evaluation on CIFAR-10 in the main submission, we analyze the standard and adversarially robust generalization gaps for RAT and the baseline defense methods on the CIFAR-100 and ImageNette datasets. 
We report the analysis results of the CIFAR-100 and ImageNette datasets in Figure~\ref{fig: gene_other}.  The results consistently demonstrate that RAT can bring smaller gaps on the standard and adversarially robust generalization, as compared to other AT-based methods. 

\begin{figure*}[ht]
\begin{center}
\includegraphics[width=0.90\textwidth]{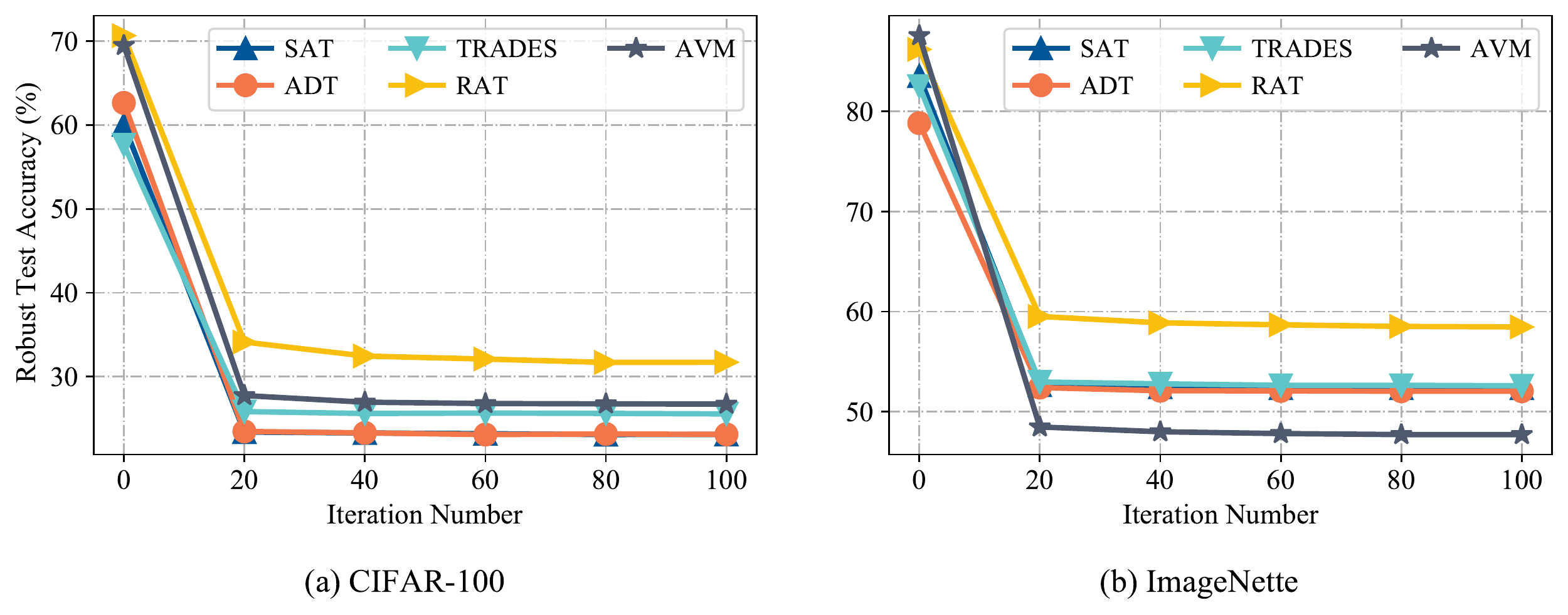} %width=0.445
\end{center}
\vspace{-0.8em}
\caption{Robust test accuracy (\%) of RAT and the baselines against PGD with various iterations on CIFAR-100 and ImageNette. The PGD perturbation budget is 8/255 and the PGD step size is 2/255.}
\label{fig: other_varying_step}
\end{figure*}

\smallskip
\textbf{Robustness against PGD with Various Iterations.} As shown in Figure~\ref{fig: other_varying_step}, we visualize the robust accuracy under the PGD attack with the iteration number $K$ varying from 0 to 100 on the CIFAR-100 and ImageNette datasets, respectively. The perturbation budget is 8/255 and the step size is 2/255. The robust accuracy of all the defense methods first decreases and then stabilizes as the iteration number increases. 
The proposed RAT consistently outperforms all the baselines under various iteration numbers, which indicates that RAT can achieve better robust generalization on the PGD attack with different number of iterations.

\begin{figure*}[ht]
\begin{center}
\includegraphics[width=0.90\textwidth]{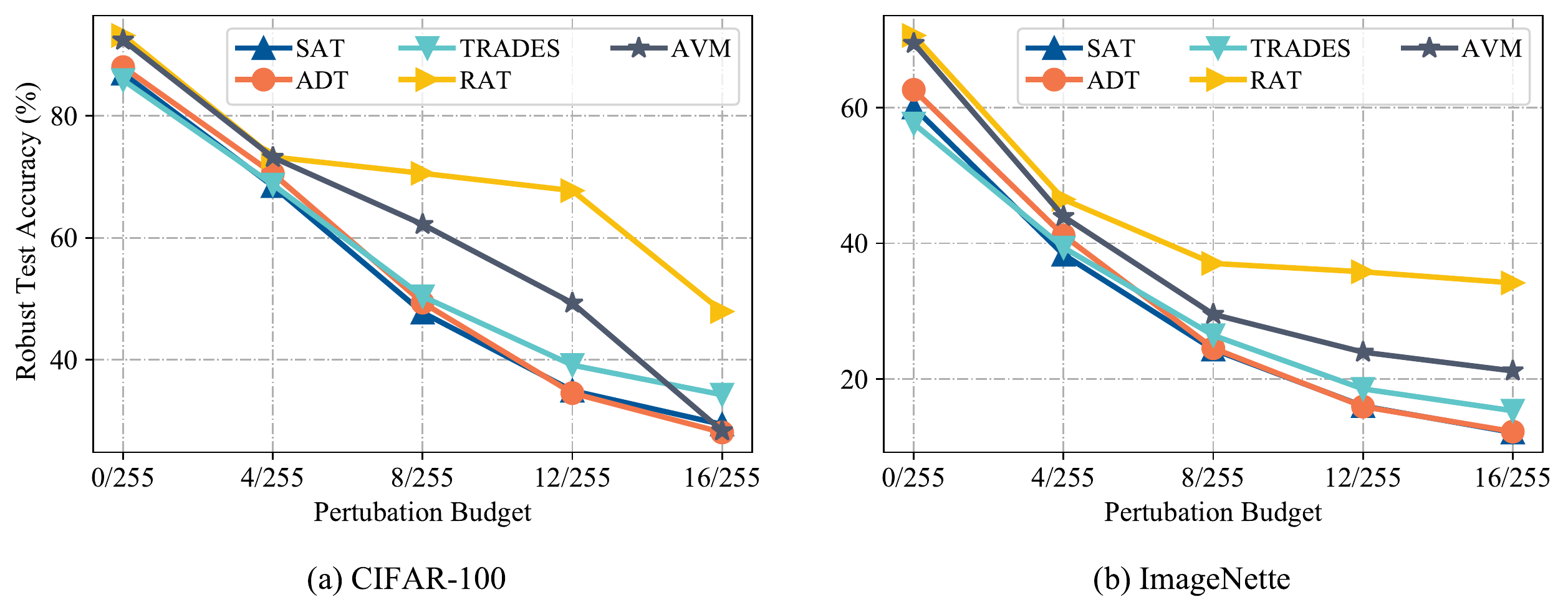} %width=0.445
\end{center}
\vspace{-0.8em}
\caption{Robust test accuracy (\%) of RAT and the baselines against PGD with various perturbation budgets on CIFAR-100 and ImageNette.
The PGD step size is 2/255 and the PGD iteration number is 10.}
\label{fig: other_varying_eps}
\end{figure*}

\smallskip
\textbf{Robustness against PGD with Various Perturbation Budgets.} In Figure~\ref{fig: other_varying_eps}, we evaluate the model robustness under the PGD attack with the perturbation budget $\epsilon$ varying from 0 to 16/255, on CIFAR-100 and ImageNette, respectively. The step size is 2/255 and the iteration number is 10. The robust accuracy of all the defense methods decreases along with the increasing of the perturbation budget. However, RAT exhibits stronger adversarial robustness, and the superiority is much clearer for higher perturbation budgets. These results consistently verify that RAT can achieve better adversarially robust generalization on the PGD attack with different perturbation budgets.

\vspace{1.2em}

\vspace{1.2em}

\end{document}